\documentclass[10pt,twocolumn,letterpaper]{article}

\usepackage[pagenumbers]{cvpr} 
\usepackage{multirow}
\usepackage{float}
\usepackage[dvipsnames]{xcolor}

\definecolor{cvprblue}{rgb}{0.21,0.49,0.74}
\usepackage[pagebackref,breaklinks,colorlinks,citecolor=cvprblue]{hyperref}

\def\paperID{*****} 
\def\confName{CVPR}
\def\confYear{2024}

\title{TCCT-Net: Two-Stream Network Architecture for Fast and Efficient Engagement Estimation via Behavioral Feature Signals}

\author{
Alexander Vedernikov\textsuperscript{1}, 
Puneet Kumar\textsuperscript{1}, 
Haoyu Chen\textsuperscript{1}, 
Tapio Seppänen\textsuperscript{1}, and 
Xiaobai Li\textsuperscript{2,1,}\thanks{Corresponding author.} \vspace{0.15cm} \\
\textsuperscript{1}{Center for Machine Vision and Signal Analysis, University of Oulu, Finland} \\
\textsuperscript{2}{State Key Laboratory of Blockchain and Data Security, Zhejiang University, China} \vspace{0.15cm} \\ 
{\tt\small \{aleksandr.vedernikov, puneet.kumar, chen.haoyu, tapio.seppanen\}@oulu.fi,} \\
{\tt\small xiaobai.li@zju.edu.cn}
}

\begin{document}
\maketitle
\begin{abstract}
Engagement analysis finds various applications in healthcare, education, advertisement, services. Deep Neural Networks, used for analysis, possess complex architecture and need large amounts of input data, computational power, inference time. These constraints challenge embedding systems into devices for real-time use. To address these limitations, we present a novel two-stream feature fusion ``Tensor-Convolution and Convolution-Transformer Network'' (TCCT-Net) architecture. To better learn the meaningful patterns in the temporal-spatial domain, we design a ``CT'' stream that integrates a hybrid convolutional-transformer. In parallel, to efficiently extract rich patterns from the temporal-frequency domain and boost processing speed, we introduce a ``TC'' stream that uses Continuous Wavelet Transform (CWT) to represent information in a 2D tensor form.
Evaluated on the EngageNet dataset, the proposed method outperforms existing baselines, utilizing only two behavioral features (head pose rotations) compared to the 98 used in baseline models. Furthermore, comparative analysis shows TCCT-Net's architecture offers an order-of-magnitude improvement in inference speed compared to state-of-the-art image-based Recurrent Neural Network (RNN) methods. The code will be released at \url{https://github.com/vedernikovphoto/TCCT_Net}. 

\end{abstract}    
\section{Introduction}
\label{sec:intro}

Real-time engagement analysis on resource-constrained mobile and embedded devices is becoming essential in the technology, education, retail, services \cite{vedernikov2024analyzing,kumar2024nontypical}. The impracticality of wired methods like electroencephalogram (EEG) or electrocardiogram (ECG) for such devices makes the utilization of facial expressions, eye gaze, head movements the superior feasible option \cite{salam202automatic,Otberdout2020Automatic,Chen2023smg}. 

The state-of-the-art (SOTA) computer vision methods for engagement analysis employ heavy image-based sequence architectures such as Recurrent Neural Networks (RNNs), Long Short-Term Memory Networks (LSTMs) \cite{selim2022students}, Temporal Convolutional Networks (TCNs) \cite{abedi2021improving, abedi2023affect}, etc. These methods demand large amount of input data (sequence of frames from videos), longer training time, significant computing resources. Yet, mobile and embedded devices require efficient algorithms for real-time engagement analysis, enhancing usability and versatility without excessive features \cite{yang2023survey}. 

\begin{figure}[!t]
  \centering
   \includegraphics[width=1\linewidth]{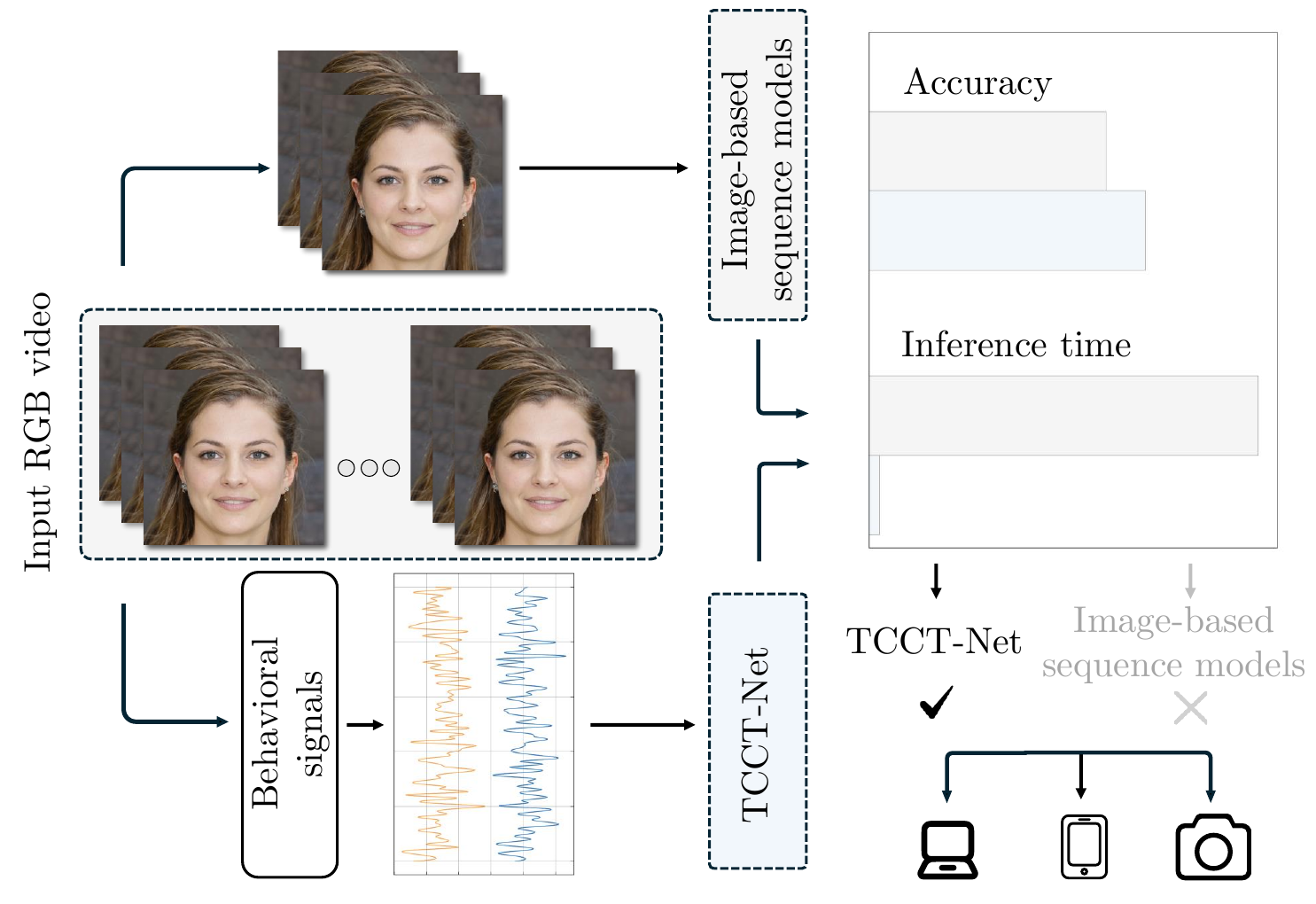}
   \caption{We introduce TCCT-Net, a novel architecture that outperforms the SOTA methods in accuracy and inference speed for the task of engagement analysis, showcasing superior efficiency and speed.}
   \label{fig:intro}
\end{figure}

Leveraging more compact signal-based input data, incorporating both temporal-spatial and frequency patterns, can address challenges in image-based architectures. Research in the affective computing field has previously utilized raw signals (making the learning low-efficient and ineffective) \cite{Li2022Regional} and RGB spectrograms processed with vision models (considered too heavy and cumbersome) \cite{Li2020Exploring,Zhang2022Spontaneous}, while insufficient attention has been paid to the utilization of 2D tensors for presenting frequency information \cite{seo20201fusing, rudd20221leveraged}. The latter balances computational demands while enriching temporal-spatial insights, enabling real-time engagement analysis in mobile and embedded computing.

Reflecting on the prior observations, we propose a lightweight yet efficient approach based on 2D tensor presentations, incorporating a two-stream network that harnesses the strengths of temporal-spatial and temporal-frequency domains for engagement analysis  (\cref{fig:intro}). The contributions of this paper are summarized as follows:
\vspace{0.1cm}

\begin{enumerate}[itemsep=3pt]
    \item We present a novel two-stream feature fusion architecture, the ``Tensor-Convolution and Convolution-Transformer Network'' (TCCT-Net). It integrates temporal-spatial-frequency data obtained from behavioral feature signals. The ``TC'' stream performs analysis on temporal-frequency behavioral features, while the ``CT'' stream focuses on their temporal-spatial analysis. TCCT-Net facilitates powerful feature extraction without complex network structures. 
    
    \item We propose a lighter alternative to conventional image-based RNNs as well as traditional temporal-frequency methods, which typically rely on raw signals and spectrograms. Specifically, we use 2D temporal-frequency tensors, derived from behavioral feature signals through Continuous Wavelet Transform (CWT).
    
    \item The proposed method demonstrates superior data learning efficiency. It needs only two behavioral feature signals as the input, unlike other methods that require dozens of behavioral feature signals or analyze dozens to hundreds of video frames.
    
    \item The TCCT-Net, evaluated on the \textit{EngageNet} dataset for predicting user engagement, outperforms existing benchmarks on both accuracy and speed while utilizing significantly fewer features.
    
\end{enumerate}
\section{Related work}
\label{sec:2_related_work}

\noindent \textbf{Automated engagement analysis.} The systematic study of automated engagement analysis by Whitehill \etal~\cite{Whitehill201486} in $2014$ marked a pivotal milestone, showcasing the potential of machine learning to estimate engagement with accuracy comparable to human judgment. Over the past decade, this field has expanded to incorporate various indicators of engagement, including features associated with facial expressions \cite{gordon2016affective}, eye gaze \cite{choi2022immersion}, body gestures \cite{khenkar2022engagement}, physiological responses \cite{shaw20221d}, textual analysis \cite{atapattu2019detecting}, reaction time \cite{ober2021detecting}, and response accuracy \cite{ober2021detecting}. Additionally, several studies have explored multi-modal approaches to further enhance the understanding of engagement \cite{filippini2021facilitating,dubovi2022cognitive}. However, not all cues or modalities are suitable for real-time engagement analysis. For example, textual analysis often involves examining tweets retrospectively \cite{hoque2022analyzing}, while the use of EEG signals \cite{shaw2023oned} and electrodermal activity (EDA) \cite{dubovi2022cognitive} is limited by the need for specialized equipment. Speech analysis may not always be relevant, particularly when participants primarily listen \cite{gupta2016daisee}. Given these constraints, the focus has shifted to methods relying solely on visual cues for engagement analysis. This eliminates the need for extra devices, streamlining the process for real-time use. 

\noindent \textbf{Vision-based engagement analysis.} Traditional computer vision techniques leverage facial expressions \cite{navarathna2017estimating}, body gestures \cite{psaltis2017multimodal}, head movements \cite{Huang2016Conversational} to automatically assess student engagement. Most SOTA computer vision approaches \cite{abedi2021improving, abedi2023affect,Savchenko20222132,selim2022students} for engagement analysis employ models that analyze frame sequences, which can range from dozens to hundreds, extracted from facial videos. The resource-intensive nature of these techniques poses challenges for real-time analysis on mobile and embedded devices. Despite their prevalence in research, their computational demands limit practical implementation. 

\noindent \textbf{Signal-based emotion analysis.} Signal-based methods offer a promising alternative. The Fast Fourier Transform (FFT), Common Spatial Pattern (CSP), Wavelet Transform (WT), and Short-Time Fourier Transform (STFT) are signal processing methods extracting frequency, spatial-frequency, or temporal-frequency features, enhancing model performance alongside temporal-spatial features \cite{xie2022transformer}. The choice of input method impacts the balance between complexity and efficiency. Overall, the signal-based methods can be categorized depending on the input data: (1) analyzing temporal-spatial patterns through time-series signal input, being too simple and ineffective, as it may not capture the nuanced features necessary for thorough analysis \cite{Li2022Regional}; (2) converting signals into images like RGB spectrograms for Convolutional Neural Network (CNN) architectures to explore temporal-frequency aspects \cite{atila2021attention,hossain2019emotion}, noting that tri-channel spectrograms might contain redundant information, which could slow down inference without offering clear benefits; and (3) emphasizing temporal-frequency features by using 2D tensors over RGB spectrograms \cite{seo20201fusing, rudd20221leveraged}. Using a single channel instead of three enhances inference speed and cuts redundant data in RGB spectrograms, yet still adding valuable temporal-frequency insights to the temporal-spatial domain. Moreover, the simplicity of 2D tensors allows for advanced on-the-fly augmentation techniques in signal processing. Overall, signal-based methods in published studies target emotion recognition using speech \cite{atila2021attention,hossain2019emotion}, EEG/EDA \cite{bota2023group}, or behavioral data \cite{singh2023do} within the temporal-spatial domain. Yet, behavioral features are unexplored in the temporal-spatial-frequency domain with 2D tensors, vital for real-time engagement analysis. Developing fast and efficient algorithms is a key for resource-constrained mobile and embedded devices.

\begin{figure*}[t]
  \centering
   \includegraphics[width=1\linewidth]{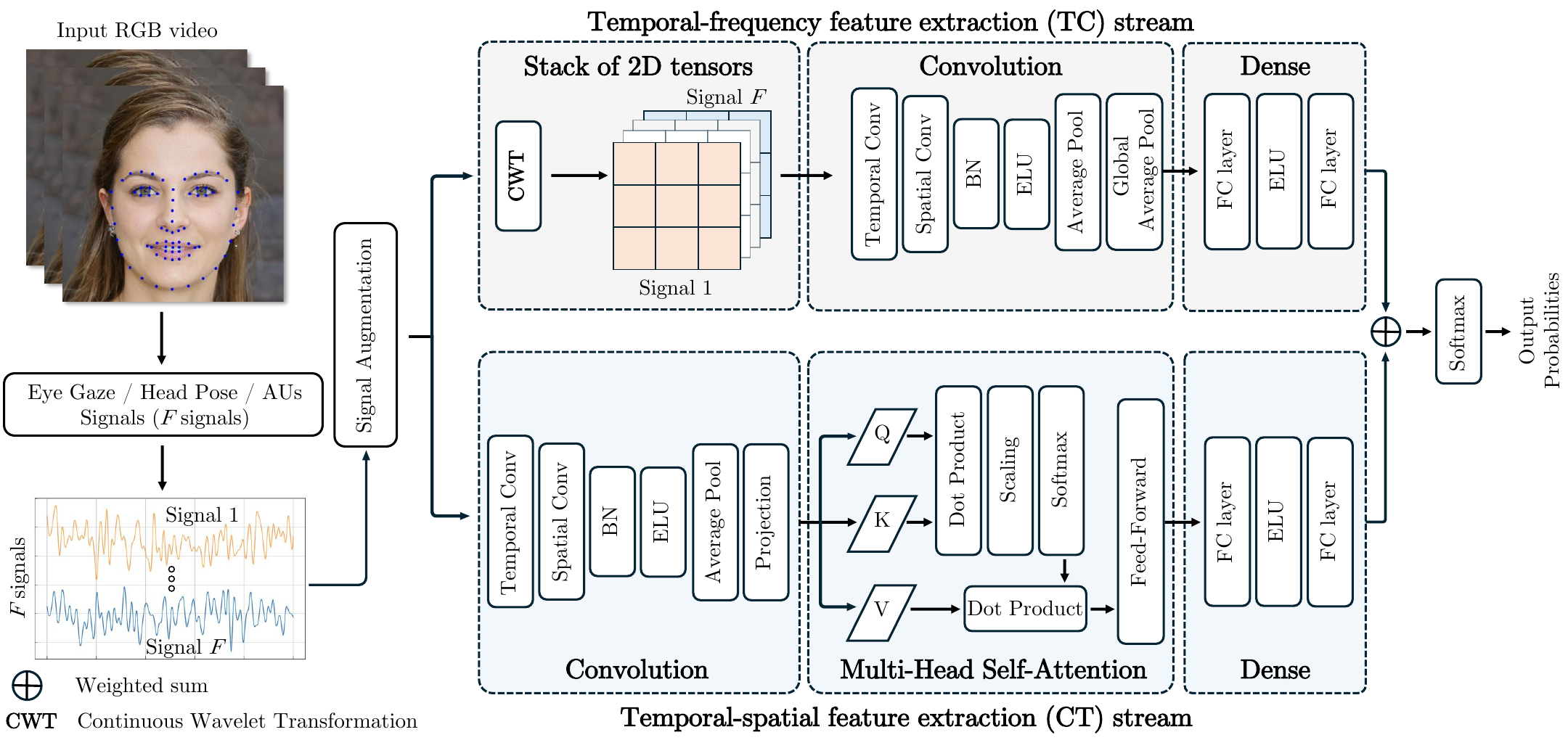}
   \caption{``Tensor-Convolution and Convolution-Transformer Network'' (TCCT-Net) architecture diagram. TCCT-Net integrates temporal-spatial-frequency data obtained from behavioral feature signals. ``TC'' stream performs analysis on temporal-frequency behavioral features, while the ``CT'' stream focuses on their temporal-spatial analysis. The predictions from both streams are fused at the decision level.}
   \label{fig:method_main}
\end{figure*}
\section{Method}
\label{sec:3_method}

\cref{fig:method_main} depicts the proposed method's framework. Initially, videos in their original RGB format undergo pre-processing, which entails analyzing each frame to extract behavioral features. Specifically, for each frame, we extract certain values corresponding to Action Units (AUs), eye gaze, and head pose. After pre-processing all the frames from the input RGB video, we obtain $F$ behavioral feature signals. Each of these signals represents a time series of values for a specific extracted feature. Stacking these signals from a video forms a two-dimensional matrix, with one dimension representing the behavioral features ($F$) and the other their temporal length. This matrix serves as the input for TCCT-Net, which comprises three main components: (1) temporal-spatial feature extraction, (2) temporal–frequency feature extraction, and (3) feature fusion and classification. This pipeline design ensures rapid model training, shorter development cycles, and achieves real-time inference performance.

\subsection{Temporal-spatial feature extraction stream}
\label{sec:ts_module}

We select the Conformer architecture \cite{song2023eeg} as a backbone of our temporal-spatial feature extraction stream due to its proven efficiency in signal data applications. We leverage Conformer's combination of convolutional layers for local pattern detection and self-attention mechanisms for understanding broad global dependencies \cite{Vaswani2017attention}. Temporal-spatial feature extraction stream begins by feeding the two-dimensional matrices (obtained during pre-processing of RGB videos) into the convolution module, where temporal and spatial convolutional layers are applied sequentially. This extracts local features and then employs average pooling for noise reduction and generalization improvement. Subsequently, the resulting representation is processed by the self-attention module, capturing long-term temporal features through global correlation analysis. Finally, a classifier with two fully connected layers outputs the decoding results for further decision-level fusion.

\noindent \textbf{Convolution module.} Building upon Schirrmeister \etal's \cite{Schirrmeister2017deep} work, the convolution module utilizes a two-step process. The first layer, with $40$ filters of dimension $1 \times 25$ and stride $1 \times 1$, focuses on extracting temporal features. The subsequent layer, with $40$ filters sized based on the number of features and a stride of $1 \times 1$, further analyzes the extracted information to identify interactions among behavioral features. For improved robustness and training efficiency, the model employs batch normalization followed by the activation function, Exponential Linear Units (ELUs) \cite{Clevert2016fast}, to introduce non-linearity. The average pooling layer (kernel size $1 \times 75$, stride $1 \times 15$) acts as a dimensionality reduction technique, optimizing feature representation. Furthermore, a dropout rate is utilized to mitigate overfitting. Before self-attention, processed features undergo transformation. Through a convolutional layer and subsequent rearrangement, a projection operation achieves this by embedding the data in a $40$-dimensional space, aligning its structure with the requirements of the following module.

\noindent \textbf{Self-Attention module.} Limited receptive fields in CNNs hinder their ability to capture long-term temporal dependencies \cite{Vaswani2017attention}. This module employs a self-attention mechanism to address this by learning global dependencies from behavioral features, complementing the previous module's limitations. To enable the self-attention mechanism, input features undergo linear transformations, generating three sets (queries (Q), keys (K), and values (V)) of equal size. The correlation between tokens is assessed through the dot product of queries and keys. A scaling factor is then applied to prevent vanishing gradients and stabilize training. Finally, a softmax operation computes attention scores used to weight the values (V) via another dot product with $d_k$ representing the dimensionality of the key vectors \cite{Vaswani2017attention}:

\vspace{-.1in}
\begin{equation}
    \text{Attention}(Q, K, V) = \text{Softmax}\left(\frac{QK^T}{\sqrt{d_k}}\right)V,
\end{equation}

To enrich representation diversity further, the model leverages a multi-head attention approach. This involves dividing the input tokens into $h$ segments, with each segment processed independently using a self-attention mechanism. Each head $l$ generates a distinct attention output based on segmented inputs $Q_l, K_l, V_l$,  which are obtained by transforming the divided tokens linearly. The resulting outputs are then concatenated, enhancing the ability of the module to capture diverse dependencies \cite{Vaswani2017attention}: 

\vspace{-.15in}
\begin{align}
    \text{MHA}(Q, K, V) &= [\text{head}_0; \ldots ; \text{head}_{h-1}], \nonumber \\
    \text{head}_l &= \text{Attention}(Q_l, K_l, V_l)
\end{align}

\noindent \textbf{Dense module.} Refinement of feature representations is achieved through a dense module employed after the multi-head self-attention stage. Linear transformations and the ELU activation function are utilized within this module, enabling the identification of intricate patterns. Dropout regularization is applied to prevent overfitting and enhance generalizability. The output of this stream, intended for classifying four distinct classes, is to be combined with the temporal-frequency stream's output via weighted decision-level fusion. 

\subsection{Temporal-frequency feature extraction stream}

By leveraging the CWT and CNN, the temporal-frequency feature extraction stream is adept at uncovering and isolating features that encompass both the temporal and spectral patterns of the behavioral signals. This enables the stream to capture a more nuanced representation of engagement, facilitating a comprehensive analysis of dynamic behaviors and patterns within videos. 
\vspace{0.1cm}

\noindent \textbf{Temporal-frequency transformation module.} A batch of behavioral feature signals is fed into the temporal-frequency transformation module. Each signal undergoes a transformation from the time domain to the frequency domain using the CWT. The CWT decomposes a signal into wavelets, allowing for the examination of different frequency components at varying scales, using the following equation \cite{rioul1992fast}:\vspace{-0.25cm}

\begin{equation}
CWT(s, \tau) = \frac{1}{\sqrt{|s|}} \int_{-\infty}^{\infty} x(t) \psi^* \left( \frac{t-\tau}{s} \right) dt
\end{equation}
where \(CWT(s, \tau)\) represents the wavelet coefficients at scale \(s\) and shift \(\tau\), \(x(t)\) is the signal to be transformed, \(\psi^*(t)\) denotes the complex conjugate of the mother wavelet \(\psi(t)\) used in the transformation. The scale factor \(s\) stretches or compresses the wavelet, and the translation parameter \(\tau\) shifts the wavelet in time. The variable \(t\) represents time, and the integral runs over the entire time domain of the signal, facilitating the analysis of the signal's different frequency components at various scales, thereby providing a time-frequency representation of the signal. Next, Complex Morlet wavelet \cite{nikolaou2002demodulation}, which is a complex sine wave modulated by a Gaussian window, was chosen as a mother wavelet function \(\psi(t)\) for its ability to provide a balance between time and frequency localization, and defined as: 
\begin{equation}
\psi(t) = \frac{1}{\sqrt{\pi B}} \exp\left(-\frac{t^2}{B}\right) \exp\left(2\pi iCt\right)
\end{equation}
where \( B \) is the bandwidth parameter, \( C \) is the center frequency, \( i \) represents the imaginary unit.

\vspace{0.1cm}
\noindent\textbf{2D tensor presentation.} The temporal-frequency transformation module produces 2D tensors, depicting behavioral features via wavelet coefficients. These coefficients quantify frequency components' magnitudes across predetermined scales, indicating the prominence of features at different time-frequency points. The tensor's horizontal dimension corresponds to the feature signal duration ($280$ units), while its vertical dimension controls computational efficiency. Larger tensors can hamper both training and inference speed due to higher computational requirements. In contrast to stitching RGB spectrograms on top of each other (which also might become problematic when dealing with dozens of features), stacking these 2D tensors offers a more coherent structure for subsequent convolutional processing. It allows for parallel, rather than sequential, processing of the feature maps during the feature extraction phase. This parallel processing approach facilitates more efficient data handling and extraction of frequency-domain features, integral for the next stage in TCCT-Net's temporal-frequency feature extraction stream. Efficiently handling multidimensional data and extracting key frequency features make this approach valuable for real-time analysis.
\vspace{0.1cm}

\noindent \textbf{Convolution module.} Following the temporal-frequency transformation, the extracted wavelet coefficients are fed into a CNN module to learn the patterns. This module builds upon the previously described convolution module (Section \ref{sec:ts_module}) but incorporates adjustments for time-frequency data. The module starts with a simple network: two convolutional layers with batch normalization, ELU activation, and average pooling. The first CNN layer, using kernel size ($1 \times 10$), applies padding only in time. This design extracts temporal features while preserving spatial information. Further, focusing on spatial relationships, the second convolutional layer (kernel size: $2 \times 1$, no padding) is applied. Global average pooling then captures global information by reducing spatial dimensions, followed by a fully connected layer adjusting the output to the dense block dimension.
\vspace{0.1cm}

\noindent \textbf{Dense module.} Sharing a structural similarity with the dense module outlined in Section \ref{sec:ts_module}, this module incorporates adjustments specific to time-frequency data processing. It outputs a vector sized by the number of classes (four). These class features are then to be fused with the temporal-spatial stream's output.

\subsection{Feature fusion and classification module}

After extracting features, the model utilizes decision-level fusion to combine temporal-spatial and temporal-frequency information. This approach leverages their complementarity for improved classification. Module predictions are merged using learnable weights optimized during training, allowing the model to adapt the combination strategy. The final class probabilities are obtained by summing the weighted outputs from both streams. This fusion mechanism effectively combines the modalities' strengths, resulting in more robust and accurate predictions.

A combined loss function is employed for performance optimization during model training. This function incorporates cross-entropy loss ($\mathcal{L}_{\text{CEL}}$) to measure prediction-label discrepancy and L2-norm regularization ($\mathcal{L}_{L_2}$) to penalize model complexity, preventing overfitting. The total loss function, $\mathcal{L}$, is given by:

\begin{equation}
\begin{aligned}
\mathcal{L} &= \mathcal{L}_{\text{CEL}} + \mathcal{L}_{L_2}= \\
&= -\frac{1}{N_b} \sum_{i=1}^{N_b} \sum_{c=1}^{M} y_{ic} \log(\hat{y}_{ic}) + \frac{\lambda}{N_b} \|\theta\|^2
\end{aligned}
\end{equation}
where $M$ is the number of classes, $N_b$
is the number of samples in a batch, $y_{ic}$ is a binary indicator ($0$ or $1$) if class label $c$ is the correct classification for instance $i$, $\hat{y}_{ic}$ is the predicted probability that instance $i$ belongs to class $c$, $\lambda$ is the regularization parameter controlling the penalty on model complexity by influencing the weight magnitudes, and $\|\theta\|^2$ represents the squared $L2$ norm of the weight vector.
\section{Experiments}

\subsection{Dataset}
We evaluate TCCT-Net on the \textit{EngageNet} dataset~\cite{singh2023do, dhall2023emotiw} proposed at the Ninth Emotion Recognition in the Wild Challenge (EmotiW) $2023$. 
\textit{EngageNet} is a large-scale dataset containing over $11,300$ video clips of $127$ participants ($18-37$ years old) interacting with a web-based platform in diverse environments. The participants were assigned to four different engagement levels (`Highly-Engaged', `Engaged', `Barely-Engaged', and `Not-Engaged') by multiple annotators. The dataset is split into subject-independent sets: $90$ participants for training, $11$ for validation, and $26$ for testing, resulting in $7,983$ training videos, $1,071$ validation videos, and $2,257$ test videos. 

\subsection{Preprocessing}

Most videos in \textit{EngageNet} have a frame rate of $30$ frame per second (fps), while some with a higher or lower frame rate. To ensure consistency, videos exceeding $30$ fps were downsampled to $30$ fps, while those with lower fps remained unchanged as it led to inferior results by distorting the natural temporal dynamics of the behavioral feature signals, as demonstrated by preliminary tests. Next, pre-processing is conducted to extract behavioral features using the OpenFace library \cite{baltrusaitis2018openface}. Since most videos are $10$ seconds long at $30$ fps (resulting in $300$ frames), OpenFace extracts $300$-element vectors (containing behavioral features for each frame). To maintain coherence in data processing, signals longer than $300$ frames were trimmed to $280$ elements. This threshold was selected after evaluating that it strikes a balance between maximizing data utilization, while preserving the authenticity of behavioral feature signals, especially for videos in the prevalent length range. Videos with a minimal number of frames were omitted, as duplicating them to meet the $280$-elements criterion would undermine the validity and reliability of the findings.

For evaluation, all $1,071$ validation set videos were used. Training employed $6,852$ videos from the training set. Videos with significantly lower fps, shorter duration, or both, posed a challenge. In the validation set, these videos' behavioral signals (if shorter than $280$ elements from OpenFace) were repeated until reaching $280$ elements (with any excess cut off). Similarly, some training set videos were processed this way. However, very short videos or those with very low fps in the training set were excluded from further analysis.

\subsection{Data augmentation}
\begin{figure}[ht]
  \centering
   \includegraphics[width=1\linewidth]{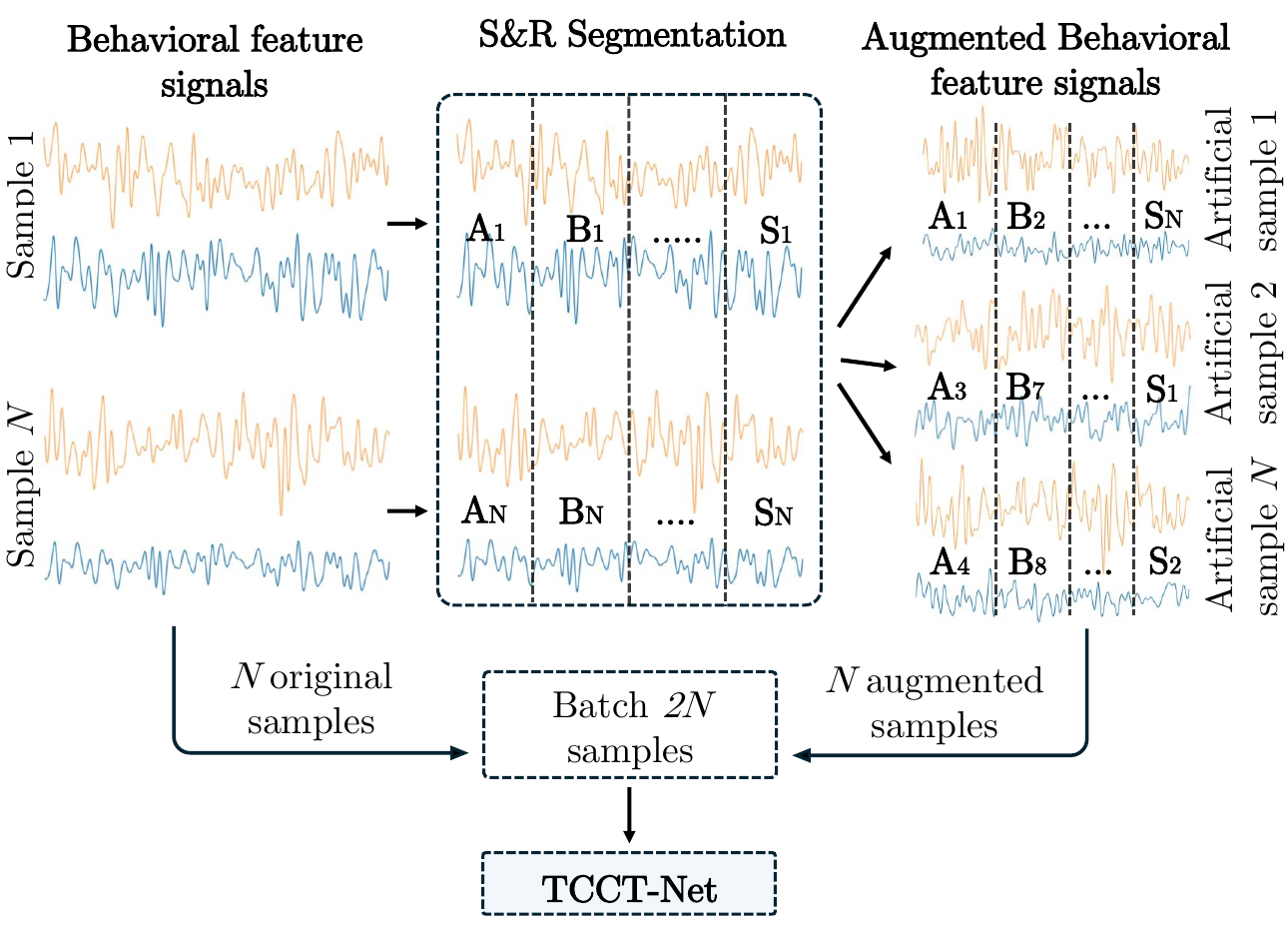}
   \caption{Segmentation and Recombination (S\&R) Augmentation. S\&R tackles overfitting by segmenting behavioral feature signal data and recombining these segments, preserving essential features while introducing realistic variations.}
   \label{fig:augm}
\end{figure}

The potential for overfitting presents a significant challenge when dealing with small signal-based datasets. Conventional methods for augmenting signal data often involve simplistic approaches like adding Gaussian noise, applying time-shifting, or altering the amplitude, either obscuring crucial features of the signal or failing to introduce meaningful diversity \cite{Lotte2018review}. These methods risk either degrading the signal quality or not fully capturing the range of variations needed to robustly train the model. In contrast, the Segmentation and Recombination (S\&R) technique \cite{Lotte2015signal}, applied in the temporal domain, effectively augments signal data by preserving its unique patterns (\cref{fig:augm}). By segmenting signals into meaningful parts and recombining them, S\&R upholds the signal's integrity and dynamics. This approach diversifies the dataset in a realistic manner without adding noise or distortions, enhancing model accuracy and generalizability.

Throughout the training process, the behavioral feature signals from the same engagement class are evenly divided into $S$ segments. These segments are then randomly concatenated. In every epoch, augmented data is generated, matching the batch size in each iteration. Consequently, the batch size utilized during training becomes $2N$, where $N$ represents the batch size specified at the data loader stage.

\subsection{Training settings}

Computational tasks were conducted on a supercomputer, utilizing $32$ SMT-enabled CPU cores and $32$ GB RAM, complemented by an AMD MI$250$ GPU with $128$GB of memory. For model training, we employed the Adam optimizer, starting with an initial learning rate of $0.0005$, and set $\beta_1 = 0.6$ and $\beta_2 = 0.999$. A learning rate scheduler was integrated to dynamically adjust the learning rate at certain epochs, optimizing the training process. During training, one batch consists of $72$ samples, each paired with additional $72$ augmented samples to enrich the dataset. To prevent overfitting and ensure efficiency, we utilized an early stopping technique, which terminated the training when further improvements in the model's performance plateaued. The exact number of epochs varied according to the specific behavioral features being utilized, as these significantly influenced the required training duration. When conducting experiments with SOTA methods, we ensured consistency by utilizing the identical computational environment.
\section{Results}
\label{sec:5_results}

In this section, we report the results of experiments involving a varying number of behavioral features, denoted as $F$ (\cref{sec:3_method}). Notably, $F=2$, representing head pose rotations around X and Y axes, yielded the highest performance. Henceforth, ``2 signals'' will denote the use of these two behavioral feature signals. Our study also included Eye Gaze ($F=2$), Facial Action Units ($F=16$) behavioral feature signals, and their fusion scenarios (\cref{subsec:5_EngageNet}).

\subsection{SOTA comparison}

In this subsection, we highlight TCCT-Net's performance against three RNN-based SOTA methods for engagement detection proposed by Abedi and Khan \cite{abedi2021improving} using ResNet + TCN, and by Selim \etal~\cite{selim2022students} employing EfficientNet B7 + LSTM and EfficientNet B7 + Bi-LSTM (\cref{tab:starting_comparison_acc}). Remarkably, TCCT-Net achieves a validation accuracy of $68.91$\% by analyzing merely $2$ signals, in contrast to RNN-based approaches, which process $50$ and $60$ frames, respectively, to achieve lower accuracies of $54.72$\%, $57.57$\%, and $58.94$\%. This focus on TCCT-Net's efficiency through minimal input signals highlights its superior performance and advancements in engagement analysis for mobile and embedded devices.

\begin{table}[h]
  \centering
  {\fontsize{9}{9.5}\selectfont
    \setlength\tabcolsep{3pt}
  \begin{tabular}{lcc}
    \toprule
    Method & Input & \begin{tabular}{@{}c@{}}Validation \\ accuracy [\%] \end{tabular} \\
    
    \midrule

    ResNet + TCN \cite{abedi2021improving} & 50 frames & 54.72 \\
    EfficientNet B7 + LSTM \cite{selim2022students} & 60 frames & 57.57 \\  
    EfficientNet B7 + Bi-LSTM \cite{selim2022students} & 60 frames & 58.94 \\  
    \underline{\textbf{TCCT-Net}} & \underline{\textbf{2 signals}} & \underline{\textbf{68.91}} \\      
    
    \bottomrule
  \end{tabular}
  
  \caption{Comparative analysis of RNN-based SOTA methods and TCCT-Net accuracy.}
  \label{tab:starting_comparison_acc}}
\end{table}

\begin{table*}[h]
  \centering
  {\fontsize{8.8}{9.5}\selectfont
  \setlength\tabcolsep{2pt} 
  \begin{tabular}{@{}lclclclclcl@{}}
    \toprule
    \multicolumn{2}{c}{EG + HP} & \multicolumn{2}{c}{AUs} & \multicolumn{2}{c}{HP} & \multicolumn{2}{c}{EG} & \multicolumn{2}{c}{EG + HP + AUs} \\ 
    \cmidrule(r){1-2} \cmidrule(r){3-4} \cmidrule(r){5-6} \cmidrule(r){7-8} \cmidrule(r){9-10}
    Method & Acc [\%] & Method & Acc [\%] & Method & Acc [\%] & Method & Acc [\%] & Method & Acc [\%] \\ 
    
    \midrule

    Transformer (28) & 64.45 & CNN-LSTM (70) & 62.00 & LSTM (12) & 67.41 & Transformer (16) & 55.45 & Transformer (98) & 65.40 \\

    LSTM (28) & 65.17 & LSTM (70) & 62.75 & TCN (12) & 67.51 & CNN-LSTM (16) & 60.78 & LSTM (98) & 66.67 \\

    {\underline{\textbf{TCCT-Net (4)}}} & {\underline{\textbf{65.64}}} & TCN (70) & 64.24 & CNN-LSTM (12) & 67.88 & LSTM (16) & 61.34 & CNN-LSTM (98) & 67.13 \\

    & & {\underline{\textbf{TCCT-Net (16)}}} & \underline{\textbf{66.29}} & {\underline{\textbf{TCCT-Net (2)}}} & \underline{\textbf{68.91}} & TCN (16) & 63.03 & {\underline{\textbf{TCCT-Net (20)}}} & {\underline{\textbf{67.13}}} \\

    & & & & & & {\underline{\textbf{TCCT-Net (2)}}} & \underline{\textbf{64.33}} & TCN (98) & 68.72 \\

    \bottomrule
  \end{tabular}
  \caption{Comparison of validation accuracy across models and analysis of performance based on different combinations of behavioral features using the \textit{EngageNet} dataset. `EG' stands for Eye Gaze, `HP' for Head Pose, and `AUs' for Facial Action Units. The number in parentheses next to each method indicates the number of features used.} 
  \label{tab:big_comparison_table}}
\end{table*}

\subsection{EngageNet comparative analysis}
\label{subsec:5_EngageNet}

The comparative analysis of TCCT-Net's validation performance against that of other publicly available models on the \textit{EngageNet} dataset, considering various combinations of behavioral features, is detailed in \cref{tab:big_comparison_table}. The authors of the \textit{EngageNet} dataset utilize features with dimensions of $16$, $12$, and $70$ for Eye Gaze, Head Pose, and Facial Action Units, respectively. In contrast, TCCT-Net operates with a significantly reduced feature set, requiring only $2$, $2$, and $16$ features for the same categories.

The substantial reduction in feature dimensions does not compromise the effectiveness of TCCT-Net, as demonstrated by its superior performance in validation accuracy across different behavioral feature combinations. It underscores TCCT-Net's capability to extract and leverage critical information from minimal data inputs, aligning with the objectives of developing lightweight models for real-time engagement analysis on resource-constrained mobile and embedded devices. Particularly noteworthy is TCCT-Net's performance in the Eye Gaze and Head Pose (EG + HP) and Facial Action Units (AUs) categories, where it achieves validation accuracies of $65.64$\% and $66.29$\%, respectively. These outcomes are notable due to the significant feature reduction from baseline methods, yet maintaining high accuracy with less computational demand.

Moreover, the performance of TCCT-Net in analyzing individual behavioral signals—Head Pose (HP) with an accuracy of $68.91$\% and Eye Gaze (EG) with $64.33$\%—further validates the model's robustness and flexibility. This indicates that TCCT-Net is not only effective in handling multi-signal input but also excels when analyzing single types of behavioral signals, making it a versatile tool for engagement analysis. The model's architecture, which fuses temporal-spatial and temporal-frequency features, plays a critical role in this achievement, enabling it to capture nuanced patterns of engagement utilizing less data. These results demonstrate that TCCT-Net offers a balance of performance and efficiency, meeting the demands of resource-constrained mobile and embedded devices. 

\subsection{Speed performance}

Evaluation of TCCT-Net involves comparing it with SOTA methods (\cref{tab:starting_comparison_speed}), with a focus on training and inference speed for potential real-time use on mobile and embedded devices. Firstly, the comparison highlights the computational efficiency of TCCT-Net utilizing 2D tensors. With training times as low as $40.1$ seconds per epoch and inference times at $2.59$ seconds for the entire validation set, TCCT-Net demonstrates an order-of-magnitude improvement over traditional RNN-based SOTA methods that rely on more computationally intensive inputs of frames extracted from videos. This efficiency is not merely a function of reduced computational complexity but also reflects an optimized balance between the quantity and quality of input data. While models like EfficientNet B7 + LSTM/Bi-LSTM \cite{selim2022students} and ResNet + TCN \cite{abedi2021improving} process larger quantities of video frames, resulting in prolonged training and inference periods, TCCT-Net achieves competitive or superior performance metrics with significantly less input data.

\begin{table}[h]
  {\fontsize{9}{9.5}\selectfont
  \centering
  \setlength\tabcolsep{2pt}
  \begin{tabular}{lcc}
    \toprule
    Method 
    & \begin{tabular}{@{}c@{}} Input \end{tabular}
    & \begin{tabular}{@{}c@{}} Train/Inference \\ time [s] \end{tabular} \\
    
    \midrule
    
    \underline{\textbf{TCCT-Net via 2D tensors}} & \underline{\textbf{2 signals}} & \underline{\textbf{40.1 / 2.59}} \\
    ResNet + TCN \cite{abedi2021improving} & 50 frames & 730 / 61.4 \\
    TCCT-Net via RGB images & 2 signals &  952 / 129  \\
    EfficientNet B7 + LSTM \cite{selim2022students} & 60 frames  & 1030 / 79.6 \\
    EfficientNet B7 + Bi-LSTM \cite{selim2022students} & 60 frames  & 1210 / 97  \\
    \bottomrule
  \end{tabular}
  \caption{Comparative analysis of RNN-based SOTA methods and TCCT-Net speed performance: epoch training and test set inference times. The aforementioned times do not include the time required for extracting behavioral features for TCCT-Net or frames for SOTA methods.}      
  \label{tab:starting_comparison_speed}}
\end{table}

The TCCT-Net model, utilizing 2D tensors from just two signals, demonstrates improved efficiency in achieving rapid training and inference times without sacrificing accuracy. This method contrasts with traditional RNN-based SOTA approaches that depend on processing extensive video frames, which demand high computational resources and time. TCCT-Net's streamlined approach not only underscores the practicality of minimal input data for real-time analysis but also highlights an optimized balance between computational efficiency and the depth of analysis. By integrating temporal-spatial and temporal-frequency domains through a novel two-stream network, TCCT-Net sets a new benchmark for engagement analysis. Its ability to deliver high performance with reduced computational demand makes it particularly suited for mobile and embedded devices, emphasizing scalability and deployment in real-world applications where speed and efficiency are paramount.

\subsection{Ablation study}

The ablation study (\cref{tab:ablation_study}) examines the individual and collective impact of the temporal-frequency stream, the temporal-spatial stream, self-attention, and data augmentation within the TCCT-Net framework, highlighting their contributions to the model's performance.

\begin{table}[ht]
  \centering
  {\fontsize{9}{9.5}\selectfont
    \setlength\tabcolsep{3pt}
  \begin{tabular}{lc}
    \toprule
    Method 
    & \begin{tabular}{@{}c@{}}Validation accuracy [\%] \end{tabular} \\
    
    \midrule     
    Only temporal-frequency stream & 57.80 \\
    TCCT-Net w/o self-attention & 63.87 \\      
    TCCT-Net w/o augmentation & 66.20 \\
    Only temporal-spatial stream  & 67.13 \\
    \underline{\textbf{TCCT-Net }} & \underline{\textbf{68.91}} \\      
    
    \bottomrule
  \end{tabular}
  
  \caption{Ablation study. All the experiments used two behavioral feature signals as the input.}
  \label{tab:ablation_study}}
\end{table}

\noindent\textbf{Single temporal-frequency stream} alone achieves a validation accuracy of $57.80$\%, which underscores the significance of capturing temporal-frequency patterns of behavioral signals. As outlined in the manuscript, this stream leverages the CWT to decompose signals into wavelets, enabling the examination of different frequency components at varying scales. This method's ability to offer a nuanced time-frequency representation of engagement signals underpins its standalone contribution to the overall architecture. It demonstrates the critical role of frequency-domain information in identifying engagement patterns, particularly when the traditional temporal-spatial analysis might overlook subtle yet informative frequency-based features.

\noindent\textbf{Single temporal-spatial stream}'s standalone performance, with a validation accuracy of $67.13$\%, highlights its efficacy in capturing the dynamic and spatial nuances of engagement behaviors. This stream combines convolution and self-attention mechanisms to extract both local and global features from behavioral signals. The convolution layers effectively capture local temporal and spatial features, while the self-attention mechanism extends the model's capability to understand long-term dependencies and global patterns within the data. This dual approach ensures a comprehensive analysis of temporal-spatial patterns, emphasizing the stream's substantial contribution to engagement analysis.

\noindent\textbf{Self-attention. }The diminished performance observed in TCCT-Net without self-attention ($63.87$\%) compared to the full TCCT-Net architecture underscores the self-attention mechanism's importance. By enabling the model to capture global dependencies and nuanced relationships within the data that might escape localized convolutional processes, self-attention enhances the model's predictive accuracy and depth of understanding. This is particularly vital in engagement analysis, where the subtleties of behavioral signals indicate profound differences in engagement levels.

\noindent\textbf{Data augmentation}'s role in boosting the model's performance is evident, with TCCT-Net without augmentation achieving $66.20$\% accuracy compared to the full model's $68.91$\%. The employment of the Segmentation and Recombination technique enriches the training dataset, ensuring robustness against overfitting and improving the generalization capabilities. This augmentation strategy further enhances the model's accuracy and its ability to adapt to new, unseen data.

The analysis of the independent contributions of the temporal-frequency and temporal-spatial streams, along with the vital roles of self-attention and data augmentation, offers a comprehensive understanding of their importance in the TCCT-Net framework. These components, individually and collectively, contribute to the model's superior performance, showcasing the sophisticated balance between depth of analysis and computational efficiency required for real-time engagement analysis.
\section{Conclusion}

We introduce TCCT-Net designed for fast and efficient engagement analysis, aimed for real-time application on resource-constrained mobile and embedded devices. This method leverages a two-stream network, integrating temporal-spatial and temporal-frequency feature extraction. TCCT-Net's demonstrated superior performance in validation accuracy while significantly reducing the computational overhead and feature input dimensions compared to the SOTA methods. This efficiency underscores the model's ability to maintain high levels of accuracy with fewer data points, showcasing its potential for real-world applications where resources are limited. Moreover, the model's speed in training and inference times further highlights its suitability for deployment in real-time systems, offering a practical solution for engagement analysis across various sectors. This work demonstrates the possibility of surpassing the SOTA methods in accuracy and speed, while relying on only two features. Future work will expand the model to other contexts and datasets to ensure its robustness. To further improve real-world applicability, we intend to incorporate remote photoplethysmography (rPPG), a technique esteemed for its emotion detection capabilities \cite{sun2022estimating}, thereby leveraging the advantages of a multimodal architecture.

\section*{Acknowledgments}
This work was supported by the Research Council of Finland (former Academy of Finland) Academy Professor project EmotionAI (grants 336116, 345122), ICT 2023 project TrustFace (grant 345948), FARIA (The Finnish-American Research \& Innovation Accelerator) project, Infotech Oulu, the University of Oulu \& Research Council of Finland Profi 7 (grant 352788). The authors also acknowledge CSC-IT Center for Science, Finland, for providing computational resources.
{
    \small
    \bibliographystyle{ieeenat_fullname}
    \bibliography{main}
}

\end{document}